\documentclass[letter,10pt, english]{article}
\usepackage[utf8]{inputenc}
\usepackage[T1]{fontenc}
\usepackage{babel}
\usepackage{hyperref}
\usepackage[utf8]{inputenc}
\usepackage{graphicx}

\author{
  Muhammad Rahman\\
  mhdrmn19@uw.edu\\
  \and
  Sachi Figliolini\\
  sachifig@gmail.com\\
  \and
  Joyce Kim\\
  jkimy@uw.edu\\
  \and
  Eivy Cedeno\\
  eivy1234@uw.edu\\
  \and
  Charles Kleier\\
  cjkleier@uw.edu\\
  \and
  Chirag Shah\\
  chirags@uw.edu\\
  \and
  Aman Chadha\\
  hi@aman.ai\\
}

\title{Artificial Intelligence in Career Counseling: A Test Case with ResumAI}

\begin{document}
\maketitle


\begin{abstract}
The rise of artificial intelligence (AI) has led to various means of integration of AI aimed to provide efficiency in tasks, one of which is career counseling. A key part of getting a job is having a solid resume that passes through the first round of programs and recruiters. It is difficult to find good resources or schedule an appointment with a career counselor to help with editing a resume for a specific role. With the rise of ChatGPT, Bard, and several other AI chat programs it is possible to provide specific, automated feedback on various concerns to suggest places for improvement within the context of career counseling. 
    
This paper begins with a quick literature review on the ethical considerations and limitations of AI in career counseling. The authors also have created their own website service, called ResumAI, to test and review the functionality of an AI career counselor.
    
The findings of this study will contribute to the understanding of chat AI ResumAI reviewer programs and sites. The implications of the findings for the field of career counseling, AI development, and ethical practice will be discussed.
\end{abstract}


\section{Introduction}
The rise of artificial intelligence (AI) has led to increasing integration of AI in mobile and web technology aimed to enhance efficiency in tasks. The recent launch and utilization of ChatGPT,  rapidly reaching 100,000,000 users within two months of launch is evidence of the massive increase in interest in AI and its applications \cite{reed}.

An under-explored area of AI utilization is in career counseling.  As of this paper’s writing, there are currently no AI chat programs with a focus of career counseling. A key part of getting a job is having a resume resume that passes through the first round of screening. Many companies utilize AI to screen resumes, which have raised concern about bias over resumes being rejected before being seen by human eye \cite{mitigating}.

For many students, especially at less well-funded schools, it is difficult to schedule an appointment with a career counselor or receive personalized feedback to help with editing a resume for a specific role. However, when career counselors at universities and colleges are able to provide this service, students report high satisfaction \cite{fadulu}. Career counseling is a field which would benefit from increased amounts of support, and an AI chatbot would help in that regard due to 24/7 availability. We define career counseling in this paper as services and activities intended to assist individuals, to make educational, training and occupational choices and to manage their careers.\cite{oecd} 

An AI career counselor would ideally be able to provide specific, automated feedback on various concerns to suggest places for improvement. However, there are ethical concerns and gaps in human and AI career counseling that need to be examined to ensure responsible and equitable use of these technologies. 

The current literature is sparse on the considerations and limitations of AI in career counseling. In general, the most identified issues of AI usage include justice, fairness, transparency, and non-maleficence, responsibility and privacy which would also apply to AI career counseling \cite{nature}.

This research paper will investigate the literature about the current usage and explorations of AI career counselors within the context of a chat AI ResumAI reviewer service. In addition to identifying select relevant papers in the literature review about this topic, the authors of the paper have also created their own service called {\em ResumeAI} to retain user feedback on their experience and goals of the usage of such a tool.


\section{Literature Review}
Using Google Scholar and PubMed, we searched relevant key terms for AI and career counseling for the  literature review. Our aim was not to be fully all-encompassing but rather to achieve a high-level overview of the state of affairs for AI, ethics, and career counseling.


A study from 2021 utilizing AI and real time data to predict with 76\% accuracy occupational transitions to help people find new jobs in times of major labor demand shifts \cite{dawson}.

Another paper reported creating a framework based on explainable AI to analyze educational factors such as aptitude, skills, expectations and educational interests to help students opt for right decisions for career growth, including choosing the correct classes. The authors used White and Black box model techniques for explainability and interpreting AI results on their career counseling-based dataset \cite{explainable}.

An AI chatbot was proposed and tested in study in 2020, allowing students confused about what career to choose to answer questions and receive feedback on their potential career pathways \cite{online}. A similar study found AI and human counselors generally agreed on recommendations, but on disagreements, the AI performed better than the prediction made by counselors \cite{efficiency}.

A study from Finland published in 2021 investigating AI in career guidance in higher education found, based on their literature review, AI being beneficial for student self-regulation, motivation and well-being as well as personalized learning support and feedback \cite{artificial}. Their own research found that students desired timely and accessible guidance, whether it be AI or human. They also found that students wanted to utilize AI to compare their skills to the requirements of specific positions.  Guidance staff themselves expressed hope AI would help them assist in relying information to students for them to better handle case management and create relationships with students. This is where services such as ResumAI will prove useful.


\section{Methods}
The authors of the paper created a website utilizing the OpenAI API text-davinci-003 (\url{https://openai.com/blog/openai-api}) to create a career counselor chatbot website. The website aims to guide the user through the career counseling service, with specific prompts on a drop down menu the user could choose to begin their session. This was to ensure the user follows specific steps to decrease the likelihood of generic advice being given.  

The process began with following the guide on the OpenAI's website on creating a chatbot. Key steps are repeated here. First, the OpenAI Github was cloned (\url{https://github.com/openai/openai-quickstart-node}). The Github contains many packages and node modules that provided the base for our application.  We installed Node.js (\url{https://nodejs.org/en}) and used 

\begin{verbatim}
npm install
\end{verbatim}

\noindent
to install the necessary requirements.  Next our API key was added in an .env file and the app was run using 

\begin{verbatim}
npm run dev
\end{verbatim}

The main components we worked with were the generate.js file which can be found under the `API' folder and the index.js file which is located within the `Pages' folder. The generate.js file's main functionality is to connect to the API. Error handling is also done at multiple steps such as checking to ensure the OpenAI API key is configured, validating a question was entered, and ensuring a resume was inputted through pdf upload or copy pasting text. If all these conditions are met, the API call occurs, leading to a generation of a response based on the prompt. There is a function, generate Prompt function takes a question and a resume as arguments that constructs a prompt to instruct the AI to act as a career advisor named ResumAI, using the resume to personalize the answer to the question.

The index.js was modified to create the appearance of the chat page component. It was made to handle pdf parsing of uploaded resumes and sample questions were also listed as a dropdown menu for users to choose from if they did not want to type their own custom question. A download to .txt file button was also added at the end of the page to allow for downloading all the chats made by the AI and user if the user wishes to save their conversation. 

A separate Github was created for the static pages which can be found here: \url{https://github.com/Eivy1234/Resumai-site}. Pages were added for the Home Page, the How to Page, and the About Us page. The main functionality remained on the index.js file which is where the ResumAI Github (\url{https://github.com/Eivy1234/capstone/tree/master}) is located. Further components were added outside of the index.js and the generate.js such as a spinner to represent loading when the user pressed the button for submitting their question. CSS style changes were made to improve the appearance of the website. 

A visual representation of this process can be found in Figure 1. 

\begin{figure}[htp]
    \centering
    \includegraphics[width=\textwidth]{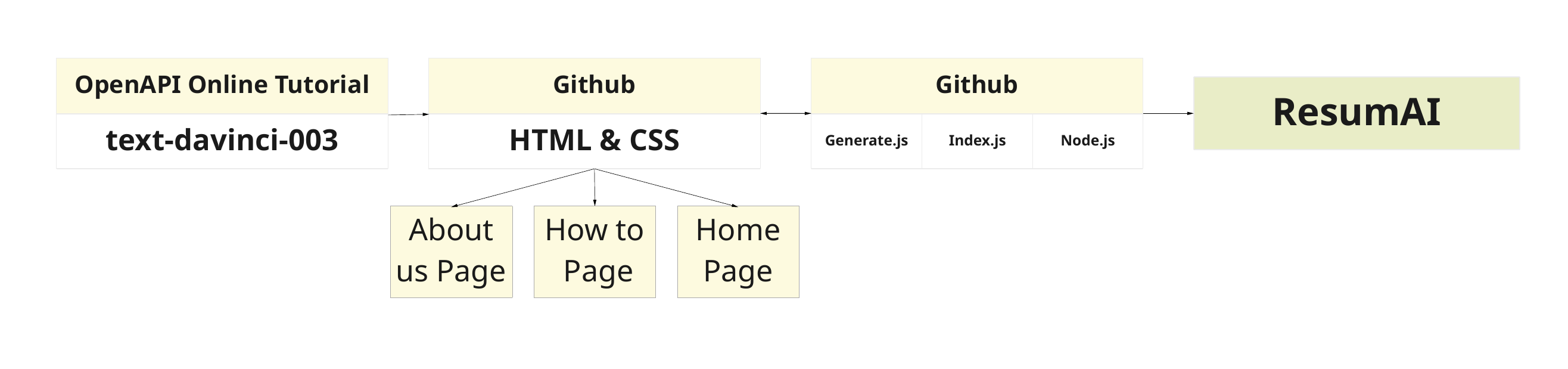}
    \caption{Architecture flow}
    \label{fig:Architecture flow}
\end{figure}


\section{Results and Discussion}
In this section we walk through a typical usage scenario with ResumAI and provide our perspectives on how it can be useful to its primary users -- students or early career professionals, especially in low-resource environments.

\subsection{ResumAI}

The ResumAI website is located at \url{https://eivy1234.github.io/Resumai-site/index.html#}\\

The following figures guides the a new user when using the service for the first time. When visiting the website service, the user is met with the home page:

\begin{figure}[htp]
    \centering
    \includegraphics[width=\textwidth]{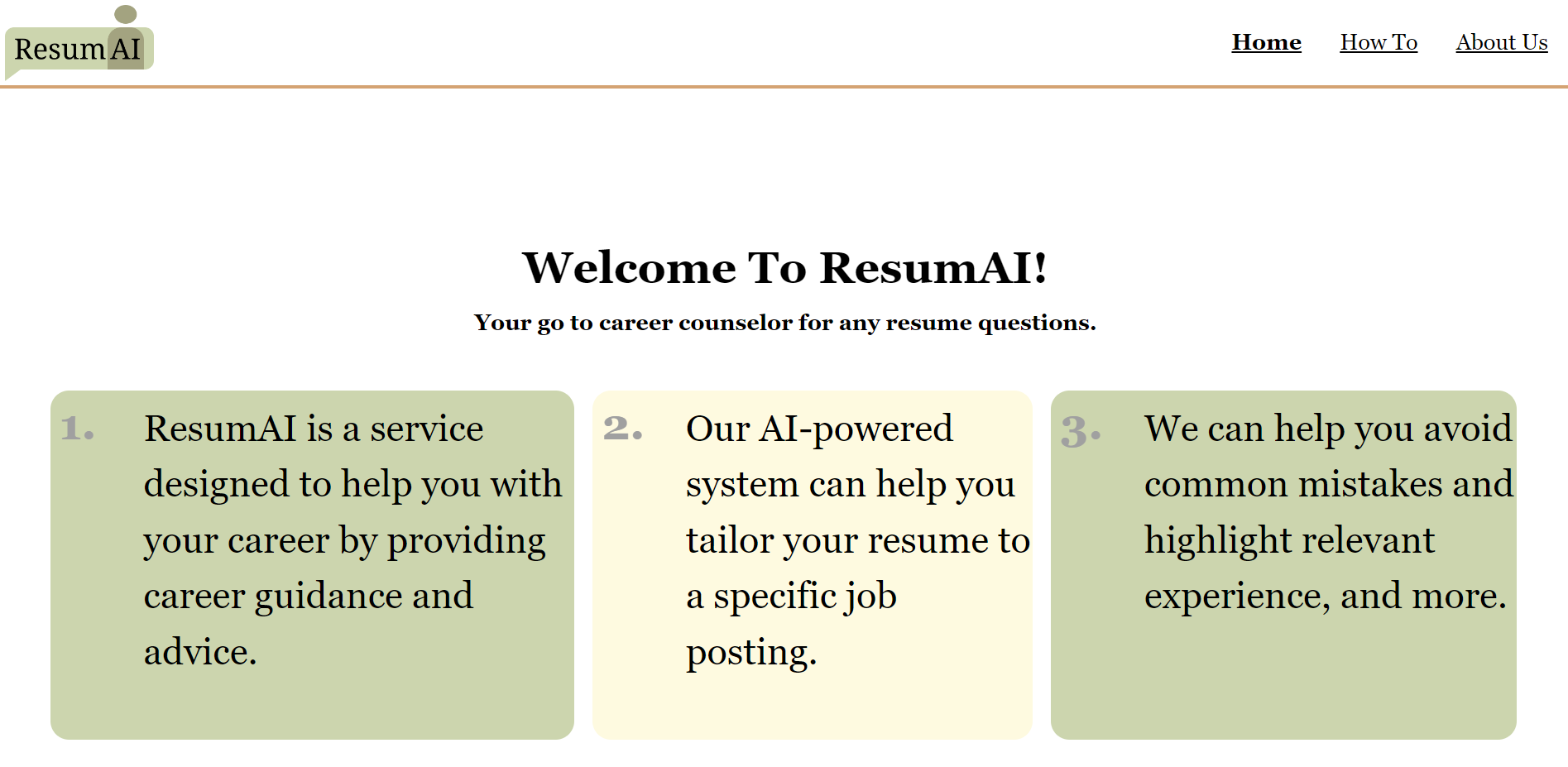}
    \caption{Step one.}
    \label{fig:Step one}
\end{figure}

\begin{figure}[htp]
    \centering
    \includegraphics[width=\textwidth]{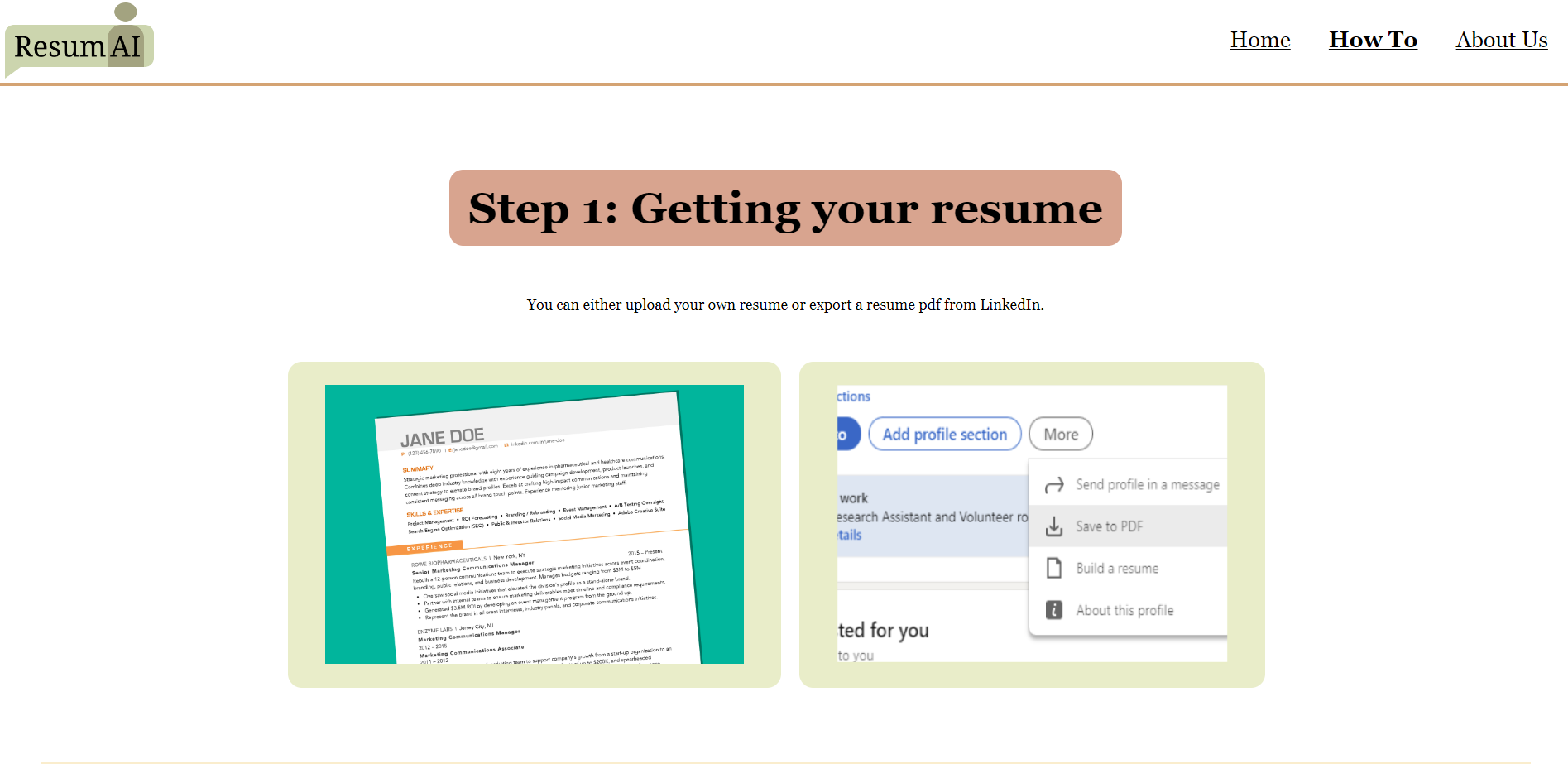}
    \caption{Step two: Clicking on the How to tab will lead them to step by step instructions.}
    \label{fig:Step two}
\end{figure}

\begin{figure}[htp]
    \centering
    \includegraphics[width=\textwidth]{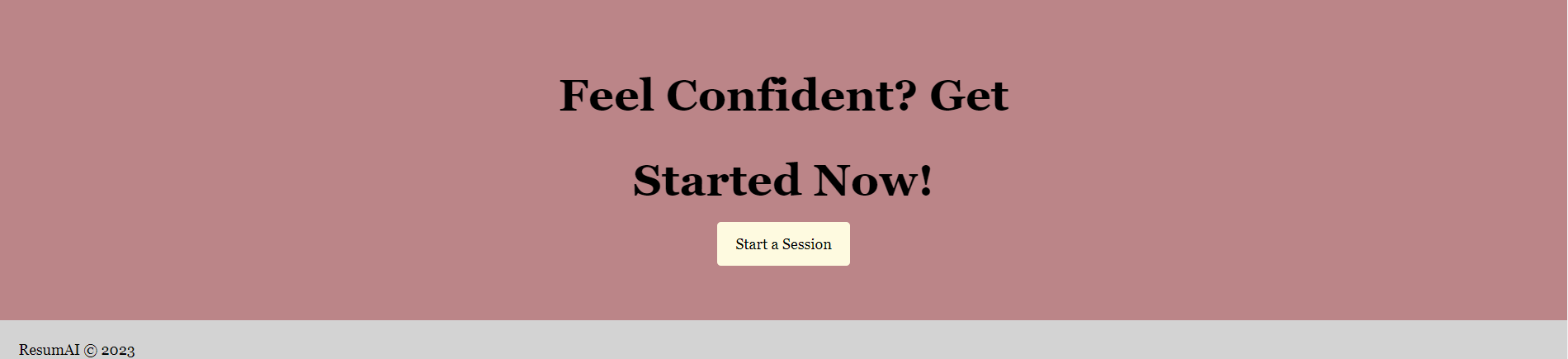}
    \caption{Step three: Once the User feels ready, they can scroll down and click the button to get started.}
    \label{fig:Step three}
\end{figure}

\begin{figure}[htp]
    \centering
    \includegraphics[width=\textwidth]{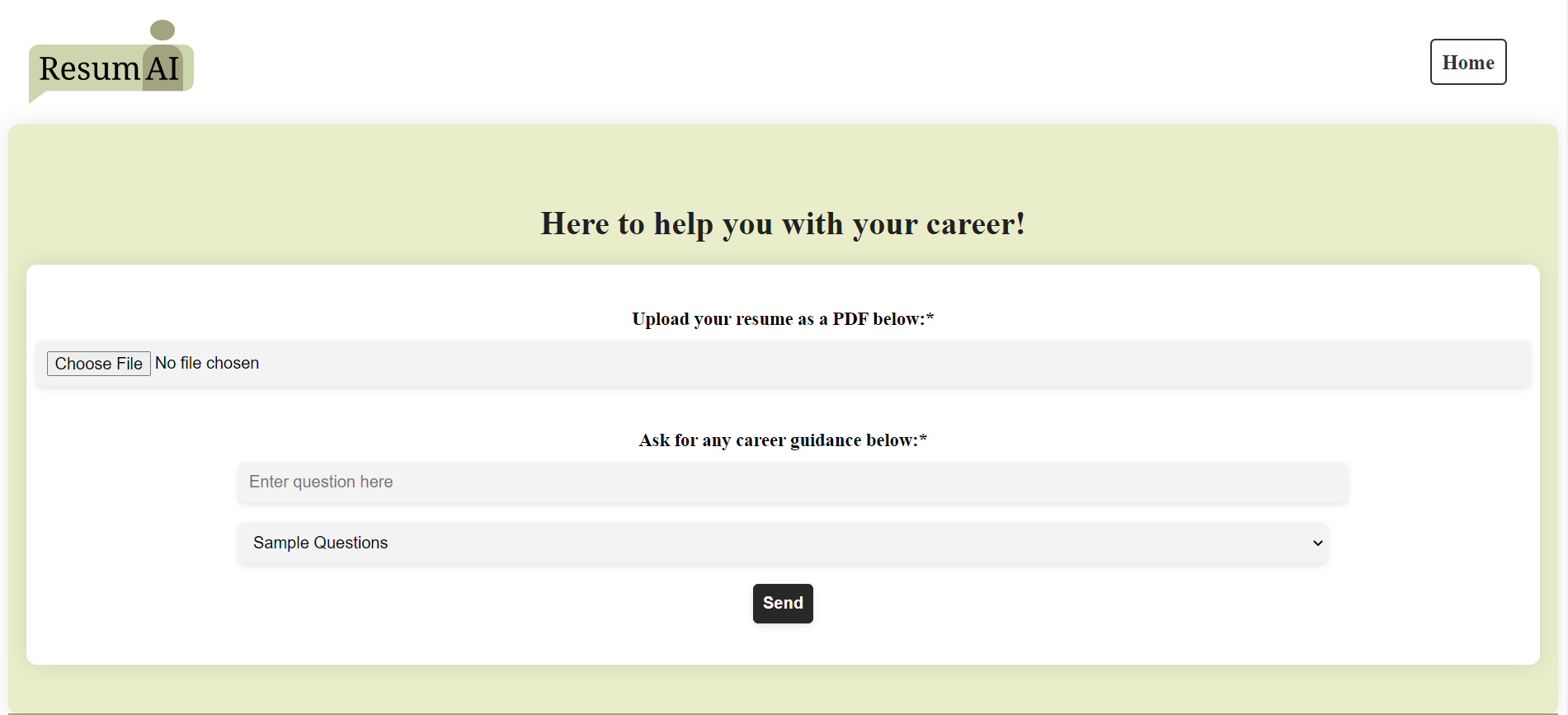}
    \caption{Once the user understands the instructions, they can feel confident to click on the Chat tab and begin their career counseling session}
    \label{fig: Chat start}
\end{figure}

We have provided a list of sample questions the user may use if they are stuck on what to ask. These include ``Are there any specific keywords or buzzwords I should include on my resume to align with industry expectations?" and ``Based on my current resume, what are the key strengths and areas for improvement?." Although surface-level, they will help new users get accustomed to using ResumAI and when they feel more prepared, they can input their custom questions at any point. Chat history is also available to be saved and downloaded into a text file in case user wants to save their conversation. 

Let us input some sample questions with a resume from an aspiring Investment Banker from the University of Pennsylvania. 

First, the student asks about his resume in a general manner. 

\begin{figure}[htp]
    \centering
    \includegraphics[width=\textwidth]{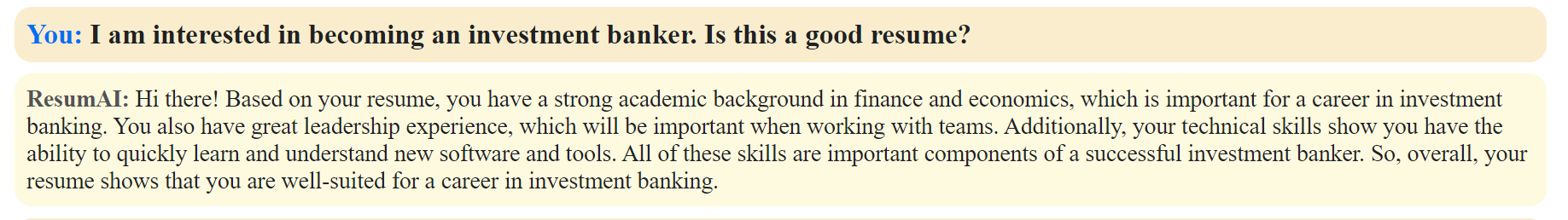}
    \caption{ResumAI on a specific career}
    \label{fig: specific career}
\end{figure}

ResumAI does a good job responding to the user prompt here. The user selected the and has a resume cleared tailed to the finance sector already and ResumAI encourages the user to continue on his career trajectory given his credentials. If he lacked the proper courses and experience, ResumAI response would have changed.

Next, the user selects the sample question of ``Are there sections or information missing from my resume that I should consider including?"

\begin{figure}[htp]
    \centering
    \includegraphics[width=\textwidth]{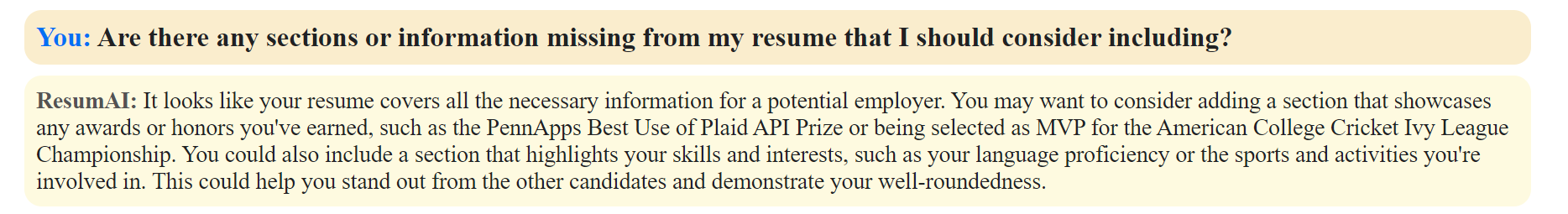}
    \caption{ResumAI on adding content on a resume}
    \label{fig: adding content}
\end{figure}

Given this prompt, ResumAI offers specific feedback on what they can improve. This quick, personalized feedback is very useful for those who are in a time-crunch or do not want to spend excessive time on formatting. The user can then go in and makes these changes and continue to ask ResumAI about the career of investment banking, his resume, or anything else related to career counseling he wishes. If he wishes to change careers for example, he could ask ResumAI what modifications he would need to achieve this career change.

We presented this service and tested with 10 undergraduate students who have all provided positive feedback. The main features they appreciated included the concise nature of the feedback as well as the ability to ask any question to ResumAI without judgment. They also appreciated how ResumAI has deep knowledge to be able to answer any question instantly on virtually any career. Otherwise they would have to set up multiple different appointments with different departments for specific expert advise on potential majors and careers. One student used it to answer her questions about the field of healthcare and engineering and potential intersections in just one session. She said it gave her ideas to consider she had not encountered before. These types of benefits were the authors' main goal when creating this service.

Limitations include that as of this paper's writing, the API utilized is unable to be fine-tuned meaning training it on custom resume datasets was not possible. As the availability of APIs increases, the next steps would be to fine-tune the models.  Furthermore, the API key we chose was the text-davinci-003. A limitation of this model is that its training data goes only until September 2021 as of July 2023, meaning it may not be able to provide the most up-to-date information about the job market.

In addition, we have included a maximum token response of 600 though this could be lifted for the ResumAI to provide even more detailed information prompt. The reasoning for this decision was to prevent ResumAI from being a one-and-one resume rewriter. The goal of the service is for career counseling to be provided, not strictly a resume service. Therefor the service will provide pointers and specific advice but will not rewrite the entire resume if prompted.

Another feature that may be added would be adding user authentication and allowing users to have cross-session chat conversations. This would allow users with very complex needs to provide and receive more information tailored to their specific situation. An advantage of this could be ResumAI's  ability to remember each user and create custom-tailored advice. The user would be able to continuously revisit the site. They could theoretically start using it as a high school student, and come back as a young professional and still receive advice on how they should navigate their careers, for example, if they wanted a career change. 

Future work includes investigating more thoroughly  the ethical concerns and gaps in human and AI career counseling to ensure responsible and equitable use of these technologies. This research paper did not entirely cover that scope with its limited literature review, so further research is needed to investigate the ethical concerns and gaps in human and AI career counseling within the context of a ResumAI reviewer-type program.


\section{Conclusion}
Utilizing AI in career counseling shows great promise in increasing accessibility for students. More research is needed on issues of potential bias and explainability of results, but for students and career changers with minimal or no career counseling available, a service like ResumAI is a much needed resource.

ResumAI exemplifies the integration of AI in career counseling through the ability to have a social impact towards low-resource job applicants and college students by supporting them through the resume development and career-finding process. With quick personalized feedback accessible to all, ResumAI provides support for those who lack resources around them to develop a strong ResumAI. ResumAI empowers individuals from low-resource and low-income backgrounds to enhance their chances of employment.

\medskip
\bibliographystyle{alpha}
\bibliography{citations}


\end{document}